\documentclass{article}
\usepackage{proceed2e}
\usepackage{amsmath}
\usepackage{amsfonts}
\usepackage{amsthm}
\usepackage{epsfig}
\usepackage{latexsym}
\usepackage{amsmath, amsthm, amssymb}

\usepackage{amsfonts}
\usepackage{amsmath}
\usepackage{amssymb}
\usepackage{latexsym}
\usepackage{color}
\usepackage{epsfig}
\usepackage{graphics}
\usepackage{enumerate}
\usepackage[sort]{natbib}

\def\Rset{\mathbb{R}}

\DeclareMathOperator*{\rank}{\mathrm rank}
\DeclareMathOperator*{\diag}{\mathrm diag}

\newcommand{\colspace}{@{\hspace{.09cm}}}
\newcommand{\bi}{\begin{itemize}}
\newcommand{\ei}{\end{itemize}}
\newcommand{\be}{\begin{enumerate}}
\newcommand{\ee}{\end{enumerate}}
\newcommand{\bd}{\begin{description}}
\newcommand{\ed}{\end{description}}
\newcommand{\ignore}[1]{}

\newcommand{\ipsfig}[2]{\scalebox{#1}{\psfig{#2}}}

\newcommand{\nys}{Nystr\"{o}m}
\providecommand{\norm}[1]{\lVert#1\rVert}
\providecommand{\minv}[1]{#1^{-1}}
\providecommand{\pinv}[1]{#1^{+}}

\newtheorem{theorem}{Theorem}

\newtheorem{definition}{Definition}

\title{Matrix Coherence and the \nys\ Method }
\author{ {\bf Ameet Talwalkar} \\  
ameet@cims.nyu.edu \\
Courant Institute of Mathematical Sciences \\  
New York, NY \\ 
\And 
{\bf Afshin Rostamizadeh}  \\ 
rostami@cims.nyu.edu \\
Courant Institute of Mathematical Sciences \\  
New York, NY \\ 
} 
 
\begin{document}

\maketitle

\begin{abstract}
The \nys\ method is an efficient technique to speed up large-scale learning
applications by generating low-rank approximations.  Crucial to the performance
of this technique is the assumption that a matrix can be well approximated by
working exclusively with a subset of its columns.  In this work we relate this
assumption to the concept of matrix coherence and connect matrix coherence to
the performance of the \nys\ method. Making use of related work in the
compressed sensing and the matrix completion literature, we derive novel
coherence-based bounds for the \nys\ method in the low-rank setting.  We then
present empirical results that corroborate these theoretical bounds.  Finally,
we present more general empirical results for the full-rank setting that
convincingly demonstrate the ability of matrix coherence to measure the degree
to which information can be extracted from a subset of columns.  
\end{abstract}	

\section{Introduction}
\label{sec:intro}
Modern problems in computer vision, natural language processing, computational
biology and other areas often involve datasets containing millions of training
instances.  However, several standard methods in machine learning, such as
spectral clustering \citep{Ng01}, manifold learning techniques
\citep{Scholkopf98,deSilva03}, kernel ridge regression \citep{krr} or other
kernel-based algorithms do not scale to such orders of magnitude.  In fact,
even storage of the matrices associated with these datasets can be problematic
since they are often not sparse and hence the number of entries is extremely
large.  As shown by \citet{Williams00}, the \nys\ method provides an
attractive solution when working with large-scale datasets by operating on only a
small part of the original matrix to generate a low-rank approximation. The
\nys\ method has been shown to work well in practice for various applications
ranging from manifold learning to image segmentation
\citep{Platt04,Fowlkes04,Talwalkar08,Zhang08}.  

The effectiveness of the \nys\ method hinges on two key assumptions on the
input matrix, $G$.  First, we assume that a low-rank approximation to $G$ can
be effective for the task at hand.  This assumption is often true empirically
as evidenced by the widespread use of singular value decomposition (SVD) and
principal component analysis (PCA) in practical applications.  As expected, the
\nys\ method is not appropriate in cases where this assumption does not hold,
which explains its poor performance in the experimental results of
\citet{Fergus09}.  Previous work analyzing the performance of the \nys\ method
incorporates this low-rank assumption into theoretical guarantees by comparing
the \nys\ approximation to the `best' low-rank approximation, i.e., the
approximation constructed from the top singular values and singular vectors of
$G$ (see Section \ref{sec:prelim} for further discussion)
\citep{Drineas05,Kumar09c}. 

The second crucial assumption of the \nys\ method involves the sampling-based
nature of the algorithm, namely that an accurate low-rank approximation can be
generated exclusively from information extracted from a small subset of $l \ll n$
columns of $G$.  This assumption is not generally true for all matrices.  
For instance, consider the extreme case of the $n \times n$ matrix described below:
\begin{equation} 
\label{eq:bad_example}
G = \left[ \begin{array}{cccccc}
\Big | & & \Big | & \Big | &  & \Big | \\ 
\vec e_1 & \ldots & \vec e_r & \vec 0 & \ldots & \vec 0 \\
\Big | & & \Big | & \Big | &  & \Big | \\ 
\end{array} \right],
\end{equation}
where $\vec e_i$ is the $i$th column of the $n$ dimensional identity matrix and
$\vec 0$ is the $n$ dimensional  zero vector.  Although this matrix has rank
$r$, it nonetheless cannot be well approximated by a random subset of $l$
columns unless this subset includes $e_1, \ldots, e_r$. In order to account for
such pathological cases, previous theoretical bounds relied on sampling columns
of $G$ from a non-uniform distribution weighted precisely by the magnitude of
the diagonal elements of $G$ \citep{Drineas05,belabbas09}.  Indeed, these
bounds give better guarantees for pathological cases.  However, in practice,
when working with real-world datasets, uniform sampling is more commonly used,
e.g., \citet{Williams00,Fowlkes04,Platt04,Talwalkar08}, since diagonal
sampling is more expensive and does not typically outperform uniform sampling
\citep{Kumar09}.  Hence the diagonal sampling bounds are not applicable in this
setting. Furthermore, these bounds are typically loose for matrices in which
the diagonal entries of the matrix are roughly of the same magnitude, as in the
case of all kernel matrices generated from RBF kernels, for which the \nys\ has
been noted to work particularly well \citep{Williams00}.

In this work, we propose to characterize the ability to extract
information from a small subset of $l$ columns using the notion of matrix
\emph{coherence}, an alternative data-dependent measurement which we believe to
be intrinsically related to the algorithm's perform. Coherence measures the
extent to which the singular vectors of a matrix are correlated with the
standard basis.  Intuitively, if we work with sufficiently incoherent matrices,
then we avoid pathological cases such as the one presented
(\ref{eq:bad_example}). Recent work on compressed sensing and matrix
completion, which also involve sampling-based approximations, have relied
heavily on coherence assumptions \citep{Donoho06, Candes06, Candes2007}.    

The main contribution of this work is the connection that is made between matrix
coherence and the \nys\ method.  Making use of related work in the compressed
sensing and the matrix completion literature, we give a more refined analysis
of this algorithm as a function of matrix coherence, presenting a novel
preliminary theoretical bound for the \nys\ method.
We also present extensive empirical results that strongly relate coherence to
the performance of the \nys\ method.   

The remainder of the paper is organized as follows. Section \ref{sec:prelim}
introduces basic definitions of coherence and gives a brief presentation of the
\nys\ method.  In Section \ref{sec:low_rank_model} we present our novel bound
for the \nys\ method under low-rank, low-coherence assumptions. Section
\ref{sec:experiments} presents extensive empirical studies that support our
bound and illustrate a similar connection between matrix coherence and the
performance of the \nys\ method for full-rank matrices.  Our empirical results
also show that incoherence assumptions are valid for several datasets derived
from real-world applications.

\section{Preliminaries}
\label{sec:prelim}
Let $G \in \Rset^{n \times n}$ be a symmetric positive semidefinite (SPSD)
matrix.  SPSD matrices, such as Gram or kernel matrices, often appear in the
context of machine learning.  For any Gram matrix, there exists an $N$ and $X
\in \Rset^{N \times n}$ such that $G = X^\top  X$.  We define $X^{(j)}, \, j =
1 \ldots n$, as the $j$th column vector of $X$ and $X_{(i)}, \, i = 1 \ldots
N$, as the $i$th row vector of $X$, and denote by $\norm{\cdot}$ the $\ell_2$
norm of a vector.  Using singular value decomposition (SVD), the Gram matrix
can be written as $G = V \Sigma V^\top $, where $V$ is orthonormal and $\Sigma
= \diag(\sigma_1, \ldots, \sigma_n)$ is a real diagonal matrix with diagonal
entries sorted in decreasing order.  For $r = \rank(G)$, the pseudo-inverse of
$G$ is defined as $G^{+} = \sum_{t=1}^r \sigma_t^{-1} V^{(t)}{V^{(t)}}^\top$.
Further, for $k \le r$, $G_k = \sum_{t=1}^k \sigma_t V^{(t)}{V^{(t)}}^\top$ is
the `best' rank-$k$ approximation to $G$, or the rank-$k$ matrix with minimal
$\norm{\cdot}_F$ distance to $G$, where $\norm{ \cdot }_F$ denotes the
Frobenius norm of a matrix.

\subsection{\nys\ method}
The \nys\ method was presented in \citet{Williams00} to speed up the
performance of kernel machines.  This is done by generating low-rank
approximations of $G$ using a subset of the columns of the matrix.
Suppose we randomly sample $l \ll n$ columns of $G$ uniformly with
replacement, and let $C$ be the $n \times l$ matrix of these sampled
columns. Then, without loss of generality, we can rearrange the
columns and rows of $G$ based on this sampling and define $X = [ X_1
\quad X_2]$ where $X_1 \in \Rset^{N \times l}$, such that  

\begin{align} 
\label{eq:blockG}
G =X^\top X & = \left[ \begin{array}{cc}
W & X_1^\top X_2 \\
X_2^\top X_1 & X_2^\top X_2 \end{array} \right]
 \\ 
\textrm{and} \quad & C = \left[ \begin{array}{c}
W \\
X_2^\top X_1
\end{array} \right], 
\end{align}
where $W = X_1^\top X_1$. The \nys\ approximation is now defined as:
\begin{equation} 
\label{eq:Nystrom}
G \approx \widetilde{G}=C \pinv{W}C^\top.
\end{equation}
The Frobenius distance between $G$ and $\widetilde{G}$, $\norm{G - \tilde{G}}_F$,
is one standard measurement of the accuracy of the \nys\ method.  The runtime
of this algorithm is O($l^3+nl^2$): $O(l^3)$ for SVD on $W$ and $O(nl^2)$ for
multiplication with $C$.  The \nys\ method is often presented with an
additional step whereby $W$ in (\ref{eq:Nystrom}) is replaced by 
\ignore{a $k$-by-$k$ (this is not true!}
its rank-$k$ approximation, $W_k$, for some
$k < l$, thus generating $\widetilde{G}_k$, the rank-$k$ \nys\ approximation to
$G$.  In this case, the runtime of the algorithm is reduced to
O($l^3+nlk$).

\subsection{Coherence}
\label{ssec:coherence}
Although the \nys\ method tends to work well in practice, the performance of
this algorithm depends on the structure of the underlying matrix. We will show
that the performance is related to the size of the entries of the singular
vectors of $G$, or the \emph{coherence} of its singular vectors.  We define $V_r$
as the top $r$ singular vectors of $G$, and denote the coherence of these
singular vectors as $\mu(V_r)$, which is adapted from \citet{Candes2007}.
\begin{definition}[Coherence]
\label{def:coherence}
The \emph{coherence} of a matrix of $V_r$ with orthonormal columns is defined
as:
\begin{equation}
\label{eq:coherence}
	\mu(V_r) = \sqrt{n} \max_{i,j} |{V_r}_{(i)}^{(j)}| \;.
\end{equation}
\end{definition}  
The coherence of $V_r$ is lower bounded by $1$, as is the case for the rank-$1$
matrix with all entries equal to $1/\sqrt{n}$, and upper bounded by $\sqrt{n}$,
as is the case for the matrix of canonical basis vectors. As discussed in
\citet{candes08,candes08b}, highly coherent matrices are difficult to randomly
recover via matrix completion algorithms, and this same logic extends to the
\nys\ method. In contrast, incoherent matrices are much easier to successfully
complete and to approximate via the \nys\ method, as discussed in Section
\ref{sec:low_rank_model}.

In order to provide some intuition, \cite{candes08} give several classes of
randomly generated matrices with low coherence.  One such class of matrices is
generated from uniform random orthonormal singular vectors and arbitrary
singular values.  For such a class they show that $\mu = O(\sqrt{\log n} \cdot
\sqrt[4]{r})$ with high probability.\footnote{For low-rank matrices,
$\sqrt[4]{r}$ is quite small.  Moreover, this $\sqrt[4]{r}$ factor only appears
due to our use of the generally loose inequality $\mu^2 \le \sqrt{r} \mu_1$,
where $\mu_1$ is a slightly different notion of coherence used in the original
bound in \citet{candes08} for this class of matrices.} In what follows, we will
show bounds on the number of points needed for reconstruction that become more
favorable as coherence decreases.  However, the bounds are useful for more
generous values of coherence than given in the above example.  We will also
provide an empirical study of coherence for various real-world and synthetic
examples.

\section{Low-rank, low-coherence bounds}
\label{sec:low_rank_model}
In this section, we make use of coherence to analyze the \nys\ method when used
with low-rank matrices.  We note that although the bounds presented throughout
this section hold for matrices of any rank $r$, they are only interesting when
$r = o(\sqrt{n})$, and hence they are most applicable in the ``low-rank''
setting.

\subsection{\nys\ method bound}
\label{ssec:nys_analysis}
The \nys\ method is empirically effective in cases where $G$ has low-rank
structure even if the matrix has full rank, i.e., $G \approx G_k$ for some $k
\ll n$.  Furthermore, as stated in Theorem \ref{thm:nys_exact} below, when $G$ is
actually a low-rank matrix, then the \nys\ method can exactly recover the
initial matrix (we include the short proof for the sake of completeness). 
\begin{theorem}[\citep{Kumar09b} Thm. 3]
\label{thm:nys_exact}
Suppose $r = \textnormal{rank}(G) \le k \le l$ and $\textnormal{rank}(W) = r$.
Then the \nys\ approximation is exact, i.e., $\norm{G - \widetilde{G}_k}_F = 0$.
\end{theorem}
\begin{proof}
Since $G = X^\top X$, $\textnormal{rank}(G) = \textnormal{rank}(X) = r$.
Similarly, $W = X_1^\top X_1$ implies $\textnormal{rank}(X_1) = r$, i.e., the
columns of $X_1$ span the columns of $X$.  We next let $U_{X_1,k}$ be the $k$
left singular vectors of $X_1$ associated with the top $k$ singular values of
$X_1$.  We then represent $W$ and $C$ in terms of $X_1$ and $X_2$, to rewrite
the \nys\ approximation as: 
\begin{align}
\widetilde{G} & = C W_k^{+}C^\top \nonumber \\
&  =  \left[ \begin{array}{c}
X_1^\top \\
X_2^\top 
\end{array} \right] X_1 (X_1^\top X_1)_k^{+} X_1^\top 
\left[ \begin{array}{cc}
X_1 &  X_2
\end{array} \right] \nonumber \\
\label{eq:nys_projection}
& = X^\top  U_{X_1,k} U_{X_1,k}^\top X.
\end{align}
Furthermore, since columns of $X_1$ span the columns of $X$,  $U_{X_1,r}$ is an
orthonormal basis for $X$ and $I - U_{X_1,r} U_{X_1,r}^\top$ is an orthogonal
projection matrix into the nullspace of $X$.\ignore{i.e., $X^\top (I - U_{X_1,r}
U_{X_1,r}^\top) \in \textnormal{Null}(X)$.} Since $k \ge r$,  from
(\ref{eq:nys_projection}) we have 
\begin{equation}
\norm{G - \widetilde{G}_k}_F = \norm{X^\top ( I - U_{X_1,k} U_{X_1,k}^\top) X}_F = 0.
\end{equation}
\end{proof}
This theorem implies that if $G$ has low-rank, then \emph{there exists} a
particular sampling such that $\rank(W) = \rank(G)$ and the \nys\ method can
perfectly recover the full matrix.  However, selecting a suitable set of $l$
columns from an $n \times n$ SPSD matrix can be an intractable combinatorial
problem, and there exist matrices for which the probability of selecting such a
subset uniformly at random is exponentially small, e.g., the rank-$r$ SPSD diagonal matrices discussed
earlier.  In contrast, a large class of SPSD matrices are much more incoherent,
and for these matrices, we will next show that by choosing $l$ to be linear in
$r$ and logarithmic in $n$ we can can with very high probability guarantee that
$\rank(W) = r$, and hence exactly recover the initial matrix. 
 
\subsubsection*{Probability of choosing a good subset}
We start with a rank-$r$ Gram matrix, $G$, and a fixed distribution,
$\mathcal{D}$, over the columns of $G$.  Our goal is to calculate the
probability of randomly choosing a subset of $l$ columns of $G$ according to
$\mathcal{D}$ such that $\rank(W)=r$.  Recall that $G=X^\top X$, $X = [ X_1
\quad X_2]$ and $W=X_1^\top X_1$. Then, by properties of SVD, we know that
$\rank(G) = \rank(X)$ and $\rank(W) = \rank(X_1)$.  Hence, the probability of this desired event is
equivalent to the probability of sampling $l$ columns of $X$ according
to $\mathcal{D}$ such that $\rank(X_1)=r$, as shown in (\ref{eq:reduce_1}).
Next, we can write the thin SVD of $X$ as $X=U_r \Sigma_r V_r^\top$, where $U_r
\in \Rset^{m \times r}, \, \Sigma_r \in \Rset^{r \times r}$ and $V_r \in
\Rset^{n \times r}$. Since $U_r$ contains orthonormal columns and $\Sigma_r$ is
invertible, we know that
\begin{equation}
\minv{\Sigma_r} U_r^\top X  =  V_r^\top. 
\end{equation}
Further, using the block representation of $X$, we have
\begin{equation}
X_1^\top U_r \minv{\Sigma_r}  =  V_{r,l}, 
\end{equation}
where $V_{r,l} \in \Rset^{l \times r}$ corresponds to the first $l$ components
for each of the $r$ right singular vectors of $X$. Since $\rank(X_1) =
\rank(X_1^\top U_r \minv{\Sigma_r})$, we obtain the equality of
(\ref{eq:reduce_2}). 
\begin{eqnarray}
\label{eq:reduce_1}
\Pr_\mathcal{D}[\rank(W) =r] & = & \Pr_\mathcal{D}[\rank(X_1) =r] \\
\label{eq:reduce_2}
& = & \Pr_\mathcal{D}[\rank(V_{r,l}) =r]. 
\end{eqnarray} 
In the next section we calculate this probability for a specific distribution
in terms of $l$ as well as a measure of the coherence of $V_{r}$.
\begin{table*}
\centering
\begin{tabular}{|\colspace l \colspace ||\colspace c \colspace |\colspace c \colspace |\colspace c \colspace |\colspace c \colspace|}
\hline
Dataset & Type of data & \# Points ($n$) & \# Features ($d$) &  Kernel  \\
\hline
PIE \citep{Sim02} & face images & $2731$ &  $2304$ & linear \\ 
MNIST \citep{mnist} & digit images & $4000$ &  $784$ & linear \\
Essential \citep{gustafson05} & proteins & $4728$ & $16$ & RBF \\
Abalone \citep{UCIdatasets} & abalones & $4177$ & $8$ & RBF \\
Dexter \citep{UCIdatasets} & bag of words & $2000$ & $20000$ &  linear \\
Artificial & random features & $1000$ & $20000$ &  linear \\
\hline
\end{tabular}
\caption{A summary of the datasets used in the experiments, including
the type of data, the number of points ($n$), the number of features ($d$) and
the choice of kernel.}
\label{table:datasets}
\end{table*}

\begin{figure*}[t]
\centering
\ipsfig{.36}{figure=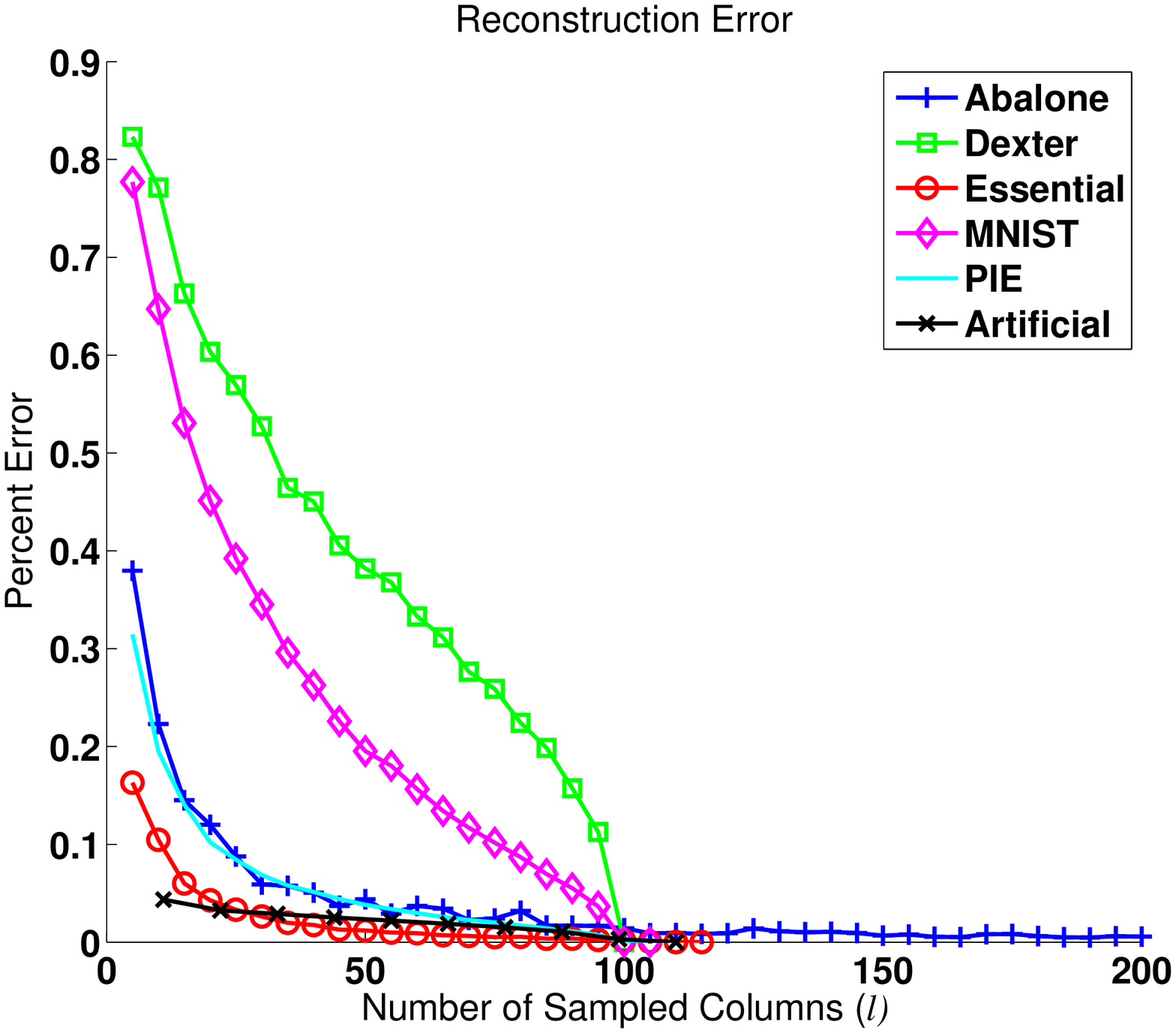}
\ipsfig{.36}{figure=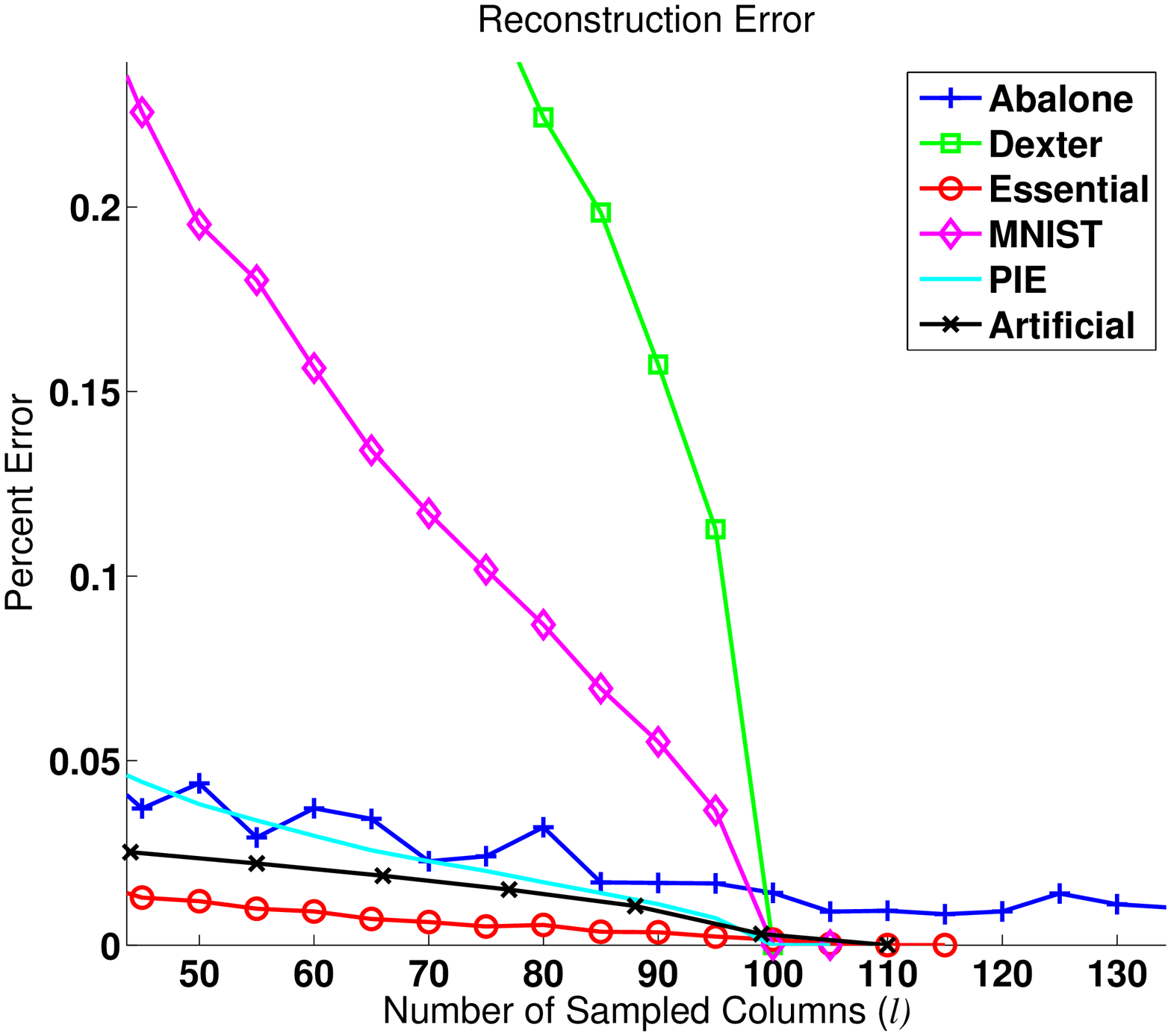}
\caption{Mean percent error over 10 trials of \nys\ approximations of rank $100$
matrices.  Left: Results for $l$ ranging from 5 to 200.  Right: Detailed view
of experimental results for $l$ ranging from 50 to 130.}
\label{fig:recon_error}
\end{figure*}

\begin{figure*}[t]
\centering
\ipsfig{.36}{figure=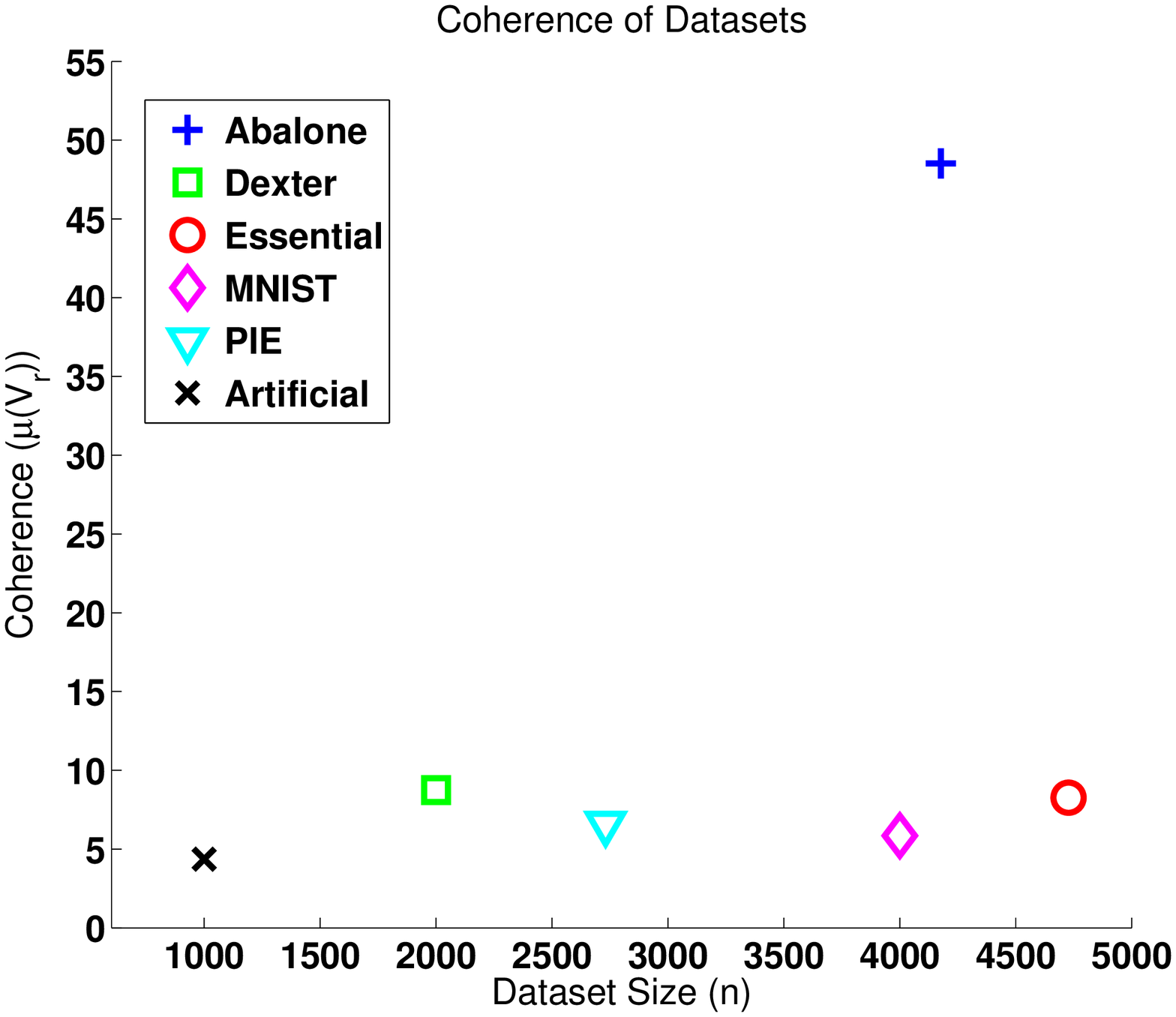}
\ipsfig{.36}{figure=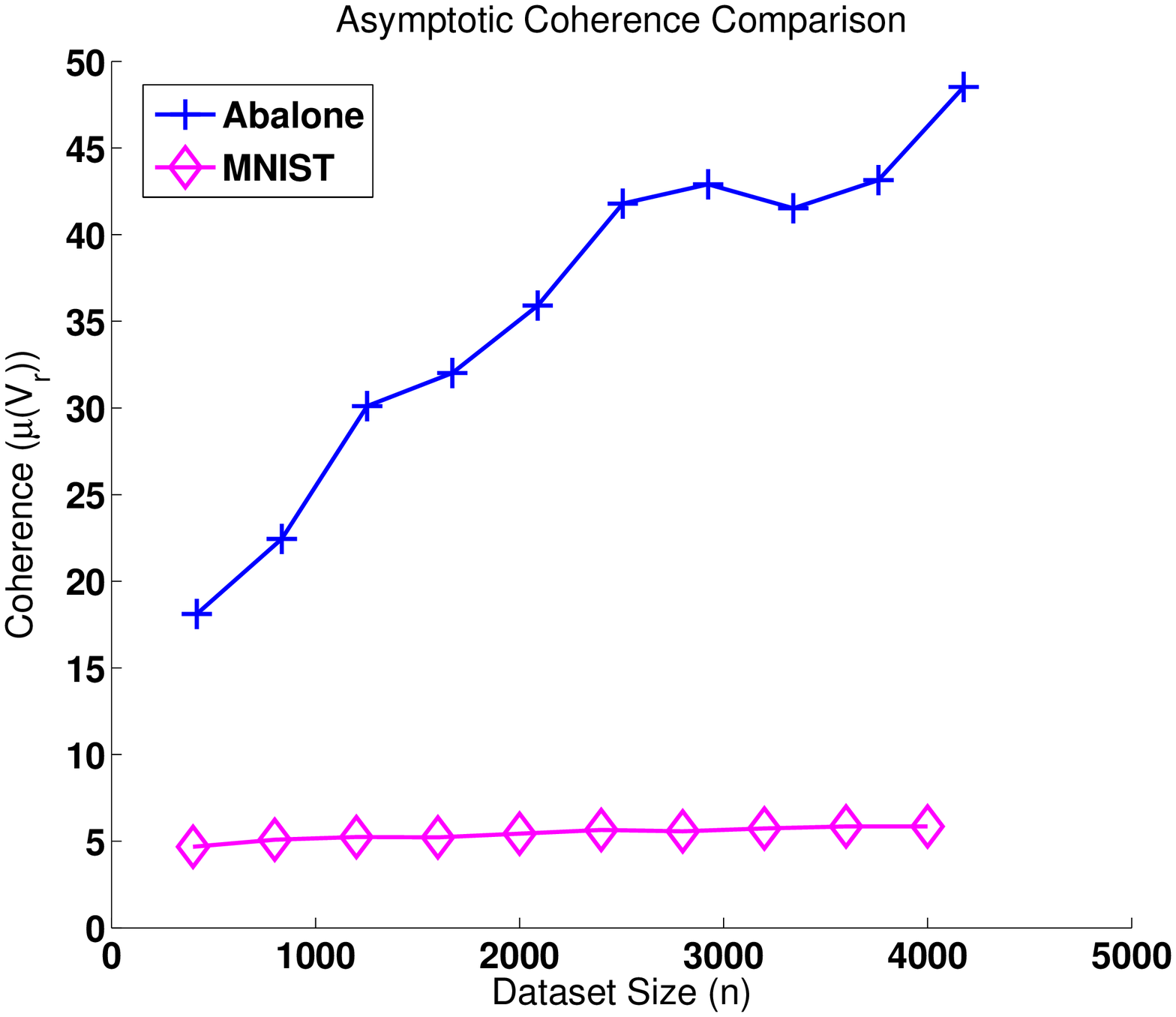}
\caption{Coherence of Datasets. Left: Coherence of rank 100 SPSD matrices used
in experiments in Section \ref{ssec:recon_experiments}.  Right: Asymptotic
growth of coherence for MNIST and Abalone datasets.  Note that coherence values
are means over ten trials.}
\label{fig:coherence}
\end{figure*}

\begin{figure*}[t]
\centering
\ipsfig{.42}{figure=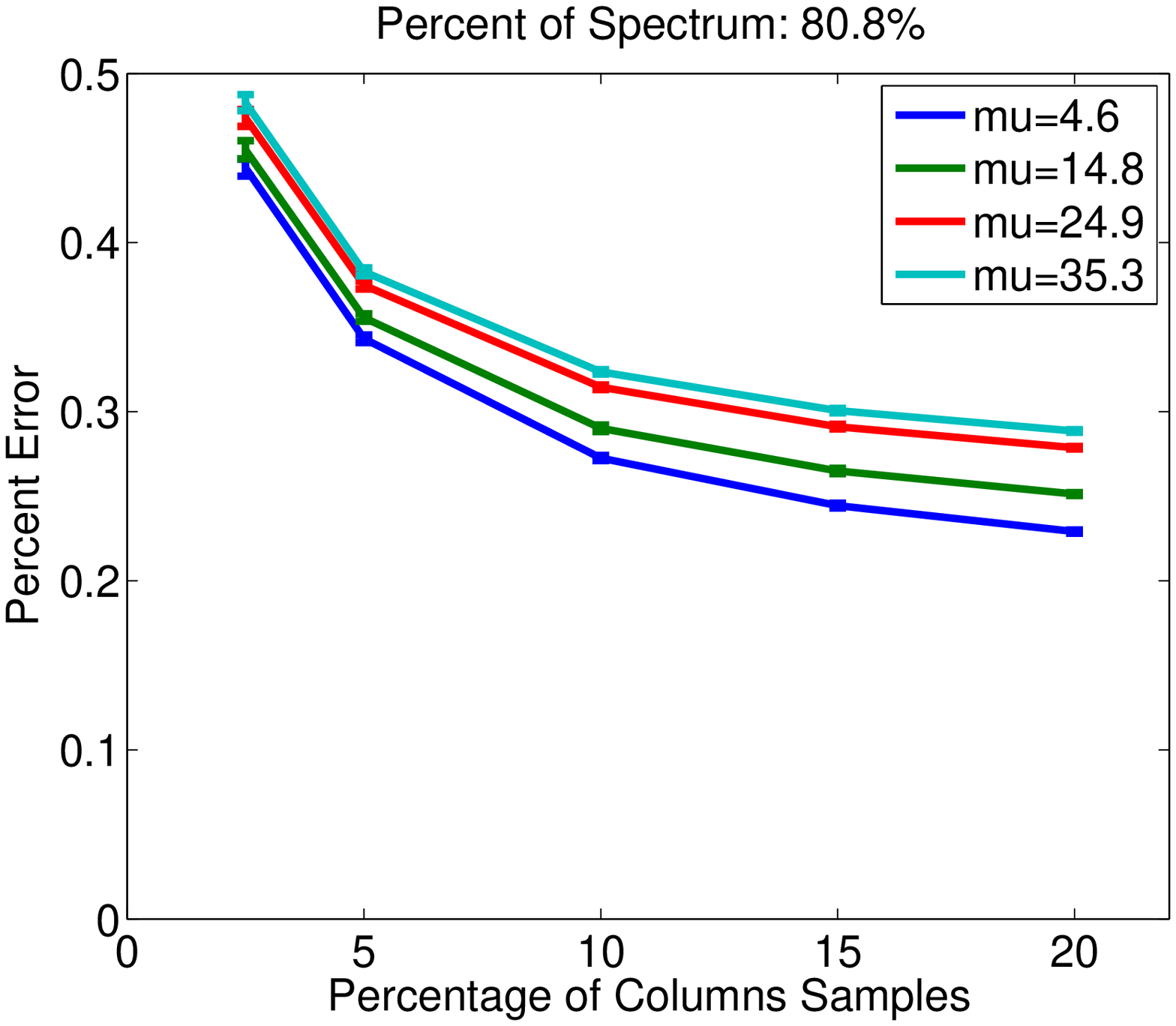}
\ipsfig{.42}{figure=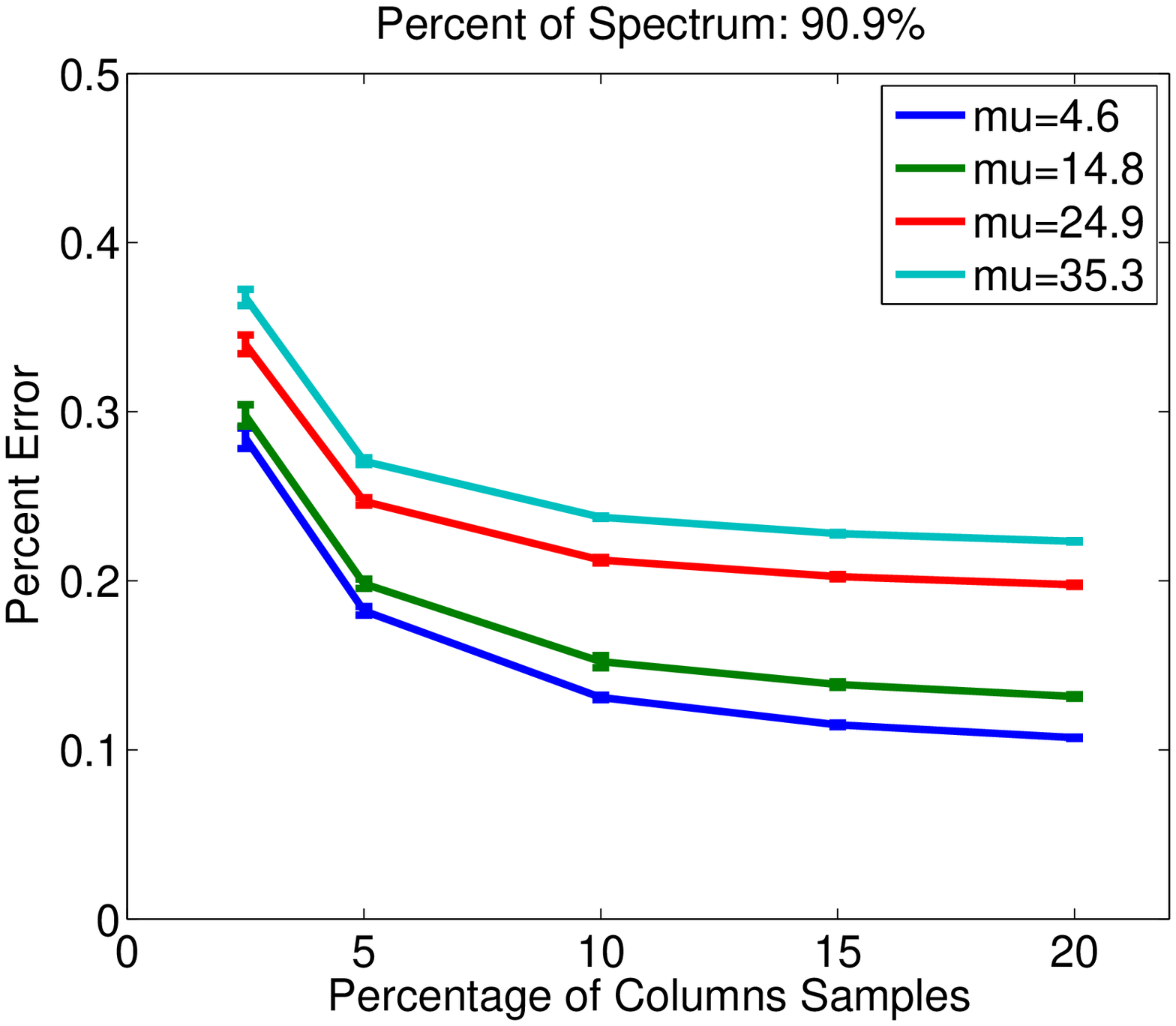}
\ipsfig{.42}{figure=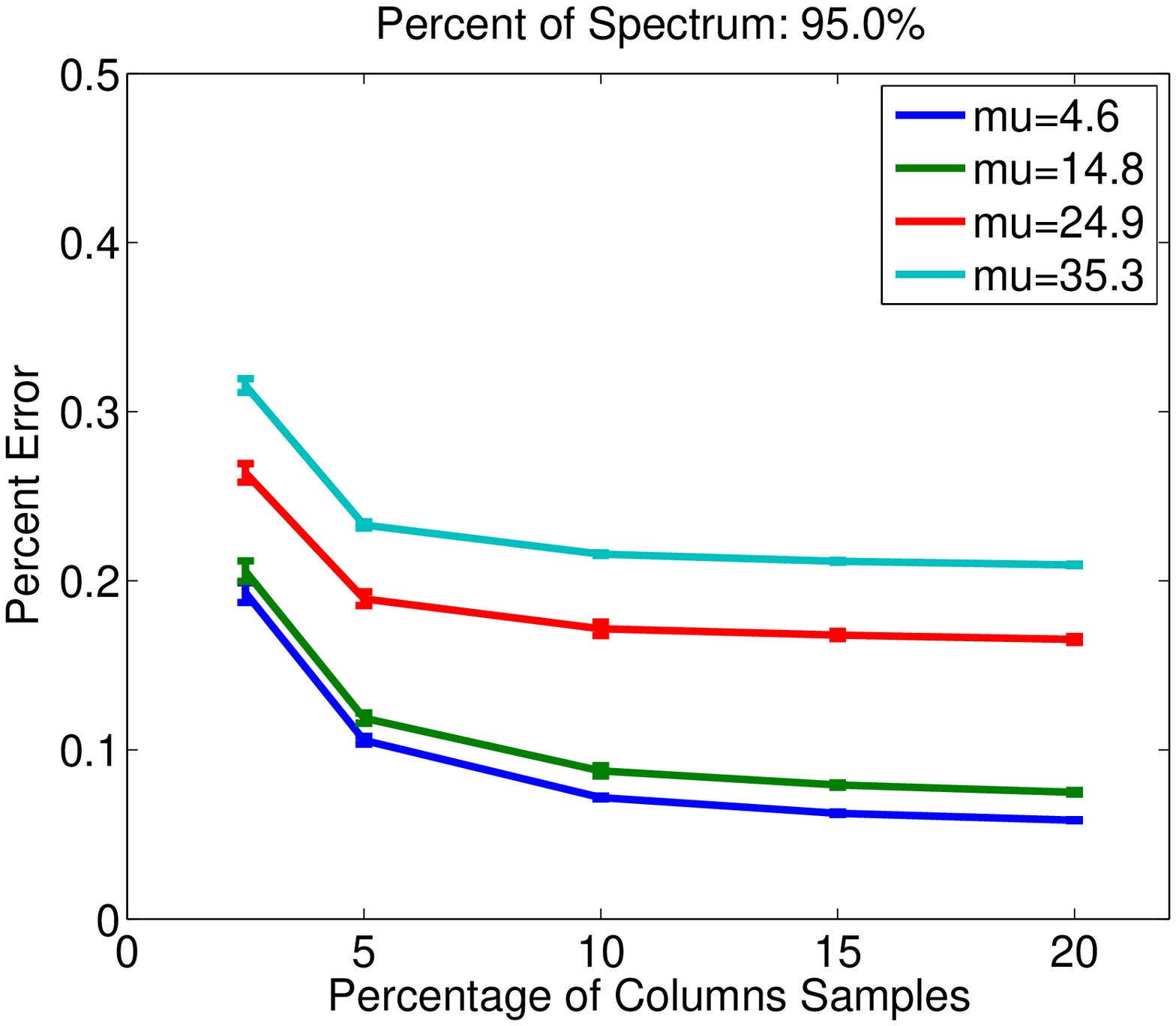}
\ipsfig{.42}{figure=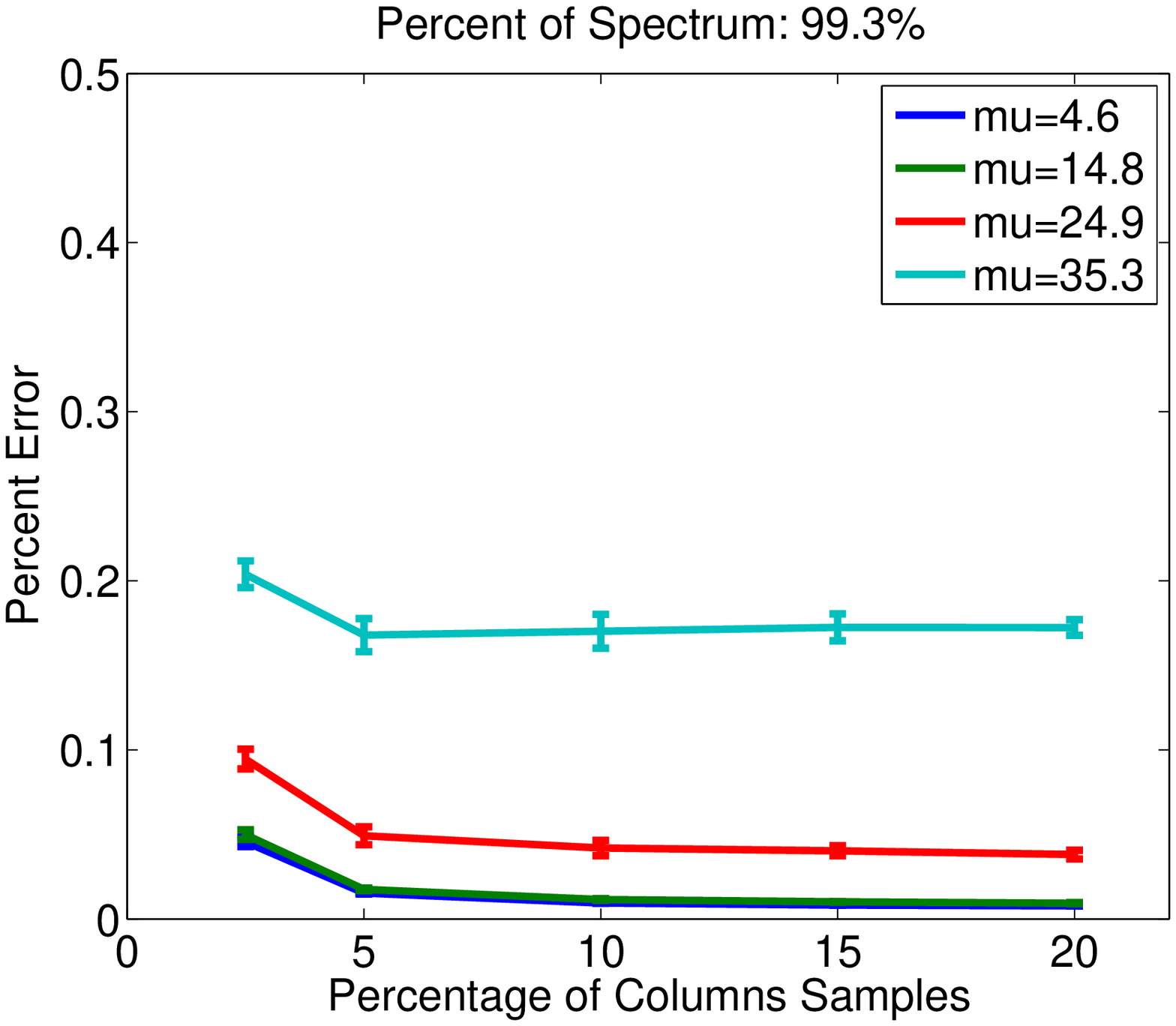}
\caption{Coherence experiments with full rank synthetic datasets, with $n=2000$
and $k=50$.  Each plot corresponds to matrices with a fixed eigenvalue decay
rate (resulting in a fixed percentage of spectrum captured) and each line
within a plot corresponds to the average results of $10$ randomly generated
matrices with the specified coherence.  Furthermore, results for each such
matrix for a fixed percentage of sampled columns are the means over $5$ random
subsets of columns. }
\label{fig:full_rank}
\end{figure*}

\subsubsection*{Sampling Bound}
Given the orthonormal matrix $V_r$, we would like to find a choice of $l$ such
that $V_{r,l}$ created by \emph{uniform} sampling has rank $r$ with high
probability. As pointed out in the previous section, a meaningful bound may not
be possible for any $l < n$ if no assumption is made on $V_{r}$.  Here we adopt
the assumption that $V_{r}$ has low coherence, as defined in Definition
\ref{def:coherence}.  We then observe that by properties of SVD we have
\begin{equation}
\label{eq:reduce_3}
\Pr \Big (\mathrm{rank}(V_{r,l}) = r \Big ) = \Pr \Big (\mathrm{rank}(V_{r,l}^\top V_{r,l}) = r \Big )\,.
\end{equation}

Next, we define $\sigma = \norm{V_{r,l}^\top V_{r,l}}_2$ and note that for $0 < c
<  1 / \sigma$, $cV_{r,l}^\top V_{r,l}$ is an $l \times l$ SPSD matrix with
singular values less than one.  Furthermore, $I - cV_{r,l}^\top V_{r,l}$ 
is also SPSD with
\begin{equation}
 \Pr \Big (\mathrm{rank}(V_{r,l}^\top V_{r,l}) = r \Big ) = \Pr \Big (\|c
V_{r,l}^\top V_{r,l} - I\| < 1 \Big ) \,,
\end{equation}
since $\|c V_{r,l}^\top V_{r,l} - I\| = 1$ implies that the nullspace of
$cV_{r,l}^\top V_{r,l}$ is nonempty.  Alternatively, if $c \ge 1 / \sigma$, then
\begin{equation}
\label{eq:reduce_4}
 \Pr \Big (\mathrm{rank}(V_{r,l}^\top V_{r,l}) = r \Big ) \ge \Pr \Big (\|c
V_{r,l}^\top V_{r,l} - I\| < 1 \Big ) \,,
\end{equation}
since, for large enough $c$, we could have $\|c V_{r,l}^\top V_{r,l} - I\| \ge
1$ even if $\mathrm{rank}(V_{r,l}^\top V_{r,l}) = r$.  Thus
the inequality in (\ref{eq:reduce_4}) holds for any constant $c > 0$, i.e.,
the probability on the RHS of (\ref{eq:reduce_4}) serves as a lower bound
for the probability of interest to us.

The probability on the RHS of (\ref{eq:reduce_4}) has been studied in previous
compressive sampling literature. Specifically, \citet{Candes2007} makes use of
a main lemma of \citet{rudelson} to derive Theorem \ref{th:candes}, which
provides us with our desired lower bound.
\begin{theorem}[ \citep{Candes2007} Thm. 1.2]
\label{th:candes}
Define $V_r \in \Rset^{n \times r}$ such that $V_r^\top V_r = I$ and let
$V_{r,l} \in \Rset^{l \times r}$ be generated from $V_r$ by sampling rows
uniformly at random. Then, the following holds with probability at least $1 -
\delta$,
\begin{equation}
	\big\| \frac{n}{l} V_{r,l}^\top V_{r,l} - I \big\|  <
\frac{1}{2} \,,
\end{equation}
for any $l$ that satisfies,
\begin{equation}
\label{eq:max}
	l \geq r \mu^2(V_r) \max \big(C_1 \log(r), C_2 \log(3 / \delta ) \big),
\end{equation}
where $C_1$ and $C_2$ are positive constants.
\end{theorem}

Note that our definition of coherence and statement of Theorem \ref{th:candes}
are modified to account for the fact that $V_r^\top V_r = I$ as oppose to $nI$,
as in \citet{Candes2007}.  Also, $V_r$ is not square as assumed in the original
theorem, however it can be verified that the proof holds even for this case.

By making use of Theorem \ref{th:candes}, we can now answer the question regarding
the number of columns needed to sample from $G$ in order to obtain an exact
reconstruction via the \nys\ method.  Theorem \ref{thm:nys_exact_incoherence}
presents a bound on $l$ for matrix completion in terms of $\mu$.
\begin{theorem}
\label{thm:nys_exact_incoherence}
Let $G \in \Rset^{n \times n}$ be a rank-$r$ SPSD matrix and assume $r
\in O(1/\delta)$, then it suffices to sample $l \geq O(r \mu^2(V_r)
\log(\delta^{-1} ) \big)$ columns to have with probability at least $1
- \delta$,
\begin{equation}
	\|G - \widetilde{G}_k \| = 0 \,.
\end{equation}
\end{theorem}
\begin{proof}
Theorem \ref{thm:nys_exact} states sufficient conditions for exact matrix
completion.  Equations (\ref{eq:reduce_1}) and (\ref{eq:reduce_2}) reduce these
sufficient conditions to a condition on the rank of $V_{r,l}$.  Equations 
(\ref{eq:reduce_3}) and (\ref{eq:reduce_4}) further reduce this problem to
a similar problem previously studied in the context of compressed sensing.
Finally, we use Theorem \ref{th:candes} to bound with high probability
the RHS of (\ref{eq:reduce_4}). 
\end{proof}
The assumption of $r \in O(1/\delta)$ is used only to simplify presentation and
avoid the appearance of a $\max$ term, as in (\ref{eq:max}). Furthermore, 
although this assumption implies that the input matrix has low rank, the low
rank setting is precisely the setting we are interested in.

\section{Experiments}
\label{sec:experiments}
In this section we present a series of empirical results that show the
empirical connection between matrix coherence and the performance of the \nys\
method.  We first perform two sets of experiments that corroborate the
theoretical claims made in the previous section --  Section
\ref{ssec:recon_experiments} illustrates the performance of the \nys\ method
for low-rank matrices using the six datasets detailed in Table
\ref{table:datasets} while Section \ref{ssec:exp_coh} interprets these results
in the context of the coherence of these datasets.  Next, we present more
general experimental results in Section \ref{ssec:exp_full_rank} that connect
matrix coherence to the \nys\ method in the case of full rank matrices.  

\subsection{Reconstruction error}
\label{ssec:recon_experiments}
In our first set of experiments we measure the accuracy of the \nys\
approximation ($\widetilde{G}_k$) for a variety of rank-$r$ matrices, with
$r=100$.  For each of the six datasets listed above, we first constructed the
optimal rank-$r$ approximation to each kernel matrix by reconstructing with the
top $r$ eigenvalues and eigenvectors.  Next, we performed the \nys\ method for
various values of $l$ to generate a series of approximations to our rank-$r$
matrix (note that we set $k=l$).  For each approximation, we calculated the
percent error of the \nys\ approximation using the notion of percent error,
defined as follows:
\begin{equation}
\text{Percent error} = \frac{\norm{G - \widetilde{G}_k}_F}{\norm{G}_F} \times 100.
\end{equation}
The results of this experiment, averaged over $10$ trials, are presented in
Figure \ref{fig:recon_error}. The figure shows that for five of the six
datasets, the \nys\ method exactly reconstructs the initial rank $r$ matrix
when the number of sampled columns ($l$) is equal or slightly larger than $r$.
Note that this observation holds for each of the ten trials, since the mean
error is zero for each of these datasets when $l \approx r$. In contrast,
for the case of the Abalone dataset, we do not see convergence to zero percent
error as $l$ surpasses $r$, and the percent error is non-zero even when $l = 2r$.

\subsection{Coherence of datasets}
\label{ssec:exp_coh}
In this set of experiments, we use the concept of coherence to explain the
results from Section \ref{ssec:recon_experiments}, namely that the \nys\ method
generates an exact matrix reconstruction for $l \approx r$ for five of the six
datasets, but fails to do so for the Abalone dataset.  As such, we first
calculated the coherence of each of the six SPSD rank $100$ matrices used in
Section \ref{ssec:recon_experiments}, using the definition of coherence from
Definition \ref{def:coherence}.  The left panel of Figure \ref{fig:coherence}
shows the coherence of these matrices with respect to the number of
points in the dataset.  This plot illustrates the stark contrast
between Abalone and the other five datasets in terms of coherence, and
helps validate our theoretical connection between low-coherence
matrices and the ability to generate exact reconstructions via the
\nys\ method.

Next, we performed an experiment in which we repeatedly subsampled the initial
SPSD matrices to generate matrices with different dimensions, i.e., different
values of $n$. For each value of $n$, we computed the coherence of the
subsampled matrix, again using Definition \ref{def:coherence}.  The right panel
of Figure \ref{fig:coherence} shows the mean results over ten trials for both
the MNIST and Abalone datasets.  As illustrated by this plot, the coherence of
the Abalone dataset grows much more quickly than that of the MNIST dataset.
As illustrated by the orthogonal random model, we expect incoherent matrices to exhibit a slow rate of growth, i.e.\ $O(\sqrt{\log n})$.  The plots for the other four
datasets (not shown) are comparable to the MNIST dataset.  These results
provide further intuition for why the \nys\ method is able to perform exact
reconstruction on all datasets except for Abalone.

\subsection{Full rank experiments }
\label{ssec:exp_full_rank}
As discussed in Section \ref{sec:intro}, the \nys\ method hinges on two
assumptions: good low-rank structure of the matrix and the ability to extract
information from a small subset of $l$ columns of the input matrix. In this
section, we analyze the effect of each of these assumptions on \nys\ method
performance on full-rank matrices, using matrix coherence as a quantification of the
latter assumption.  To do so, we devised a series of experiments using
synthetic datasets that precisely control the effects of each of these
parameters.  

To control the low-rank structure of the matrix, we generated artificial datasets with exponentially
decaying eigenvalues with differing decay rates, i.e., for $i \in \{1, \ldots,
n\}$ we defined the $i$th singular value as $\sigma_i = \exp(-i\eta)$, where
$\eta$ controls the rate of decay.  For a fixed value of $\eta$, we then
measured the percentage of the spectrum captured by the top $k$ singular values
as follows:
\begin{equation}
\textnormal{Percent of Spectrum} = \frac{\sum_{i=1}^k \sigma_{i}}{\sum_{i=1}^n \sigma_{i}}.
\end{equation}
To control coherence, we generated singular vectors with varying coherences by
forcing the first singular vector to achieve our desired coherence and then
using QR to generate a full orthogonal basis.  The smallest values of $\mu$
used in our experiments correspond to randomly generated orthogonal matrices.
We report the results of our experiments in Figure \ref{fig:full_rank}.  For
these experiments we set $n=2000$ and $k=50$. Each plot corresponds to matrices
with a fixed eigenvalue decay rate (resulting in a fixed percentage of spectrum
captured) and each line within a plot corresponds to the average results of $10$
randomly generated matrices with the specified coherence.  Furthermore, results
for each such matrix for a fixed percentage of sampled columns are the means
over $5$ random subsets of columns.   

There are two main observations to be drawn from our experiments.  First, as
noted in previous work with the \nys\ method, the \nys\ method generates better
approximations for matrices with better low rank structure, i.e., matrices with
a higher percentage of spectrum captured by the top $k$ singular values.
Second, following the same pattern as in the low-rank setting, the \nys\ method
generates better approximations for lower coherence matrices, and hence, matrix
coherence appears to effectively capture the degree to which information can be
extracted from a subset of columns. 

\section{Conclusion and future work}
In this work, we make a connection between matrix coherence and the performance
of the \nys\ method. Making use of related work in the compressed sensing and
the matrix completion literature, we derive novel coherence-based bounds for
the \nys\ method in the low-rank setting.  We then present empirical results
that corroborate these theoretical bounds.  Finally, we present more general
empirical results for the full-rank setting that convincingly demonstrate the
ability of matrix coherence to measure the degree to which information can be
extracted from a subset of columns. Future work involves  developing algorithms
to efficiently estimate the coherence of a dataset to help quickly determine
the applicability of the \nys\ method on a case-by-case basis as well as
generalizing our coherence-based bounds to the case of full rank matrices. 

\bibliographystyle{aaai-named} 
\bibliography{nys_exact} 

\end{document}